\documentclass{article}

\usepackage[letterpaper]{geometry}
\usepackage[parfill]{parskip}
\PassOptionsToPackage{numbers, compress}{natbib}
\usepackage[numbers, compress]{natbib}
\usepackage{xr-hyper,refcount}

\usepackage{graphicx}
\usepackage[utf8]{inputenc} 
\usepackage[T1]{fontenc}    
\usepackage{url}            
\usepackage{booktabs}       
\usepackage{amsfonts}       
\usepackage{nicefrac}       
\usepackage{microtype}      
\usepackage[parfill]{parskip}
\usepackage{algorithm,algorithmic}
\usepackage{amsmath,amsthm,amssymb,bbm}
\usepackage{mathtools}
\usepackage{cases}
\usepackage{comment}
\usepackage{subcaption}
\usepackage{algorithm,algorithmic}
\usepackage{color}
\usepackage{appendix}
\usepackage{xspace}
\usepackage{enumitem}
\usepackage[english]{babel}
\usepackage{authblk}

\usepackage[colorlinks,citecolor=blue,urlcolor=blue,linkcolor=blue,linktocpage=true]{hyperref}
\usepackage{cleveref}
\crefformat{equation}{(#2#1#3)}
\crefrangeformat{equation}{(#3#1#4) to~(#5#2#6)}
\crefname{equation}{}{}
\Crefname{equation}{}{}

\crefname{definition}{\textbf{definition}}{definitions}
\Crefname{definition}{Definition}{Definitions}
\crefname{assumption}{\textbf{assumption}}{assumptions}
\Crefname{assumption}{Assumption}{Assumptions}
\definecolor{maroon}{RGB}{192,80,77}

\newtheorem{theorem}{Theorem}
\newtheorem{lemma}[theorem]{Lemma}

\newtheorem{corollary}[theorem]{Corollary}

\newtheorem{remark}[theorem]{Remark}

\usepackage[utf8]{inputenc} 
\usepackage[T1]{fontenc}    
\usepackage{hyperref}       
\usepackage{url}            
\usepackage{booktabs}       
\usepackage{amsfonts}       
\usepackage{nicefrac}       
\usepackage{microtype}      

\usepackage{natbib}
\usepackage{amsmath}
\usepackage{bbm}

\usepackage[normalem]{ulem}

\usepackage{mathtools}

\DeclarePairedDelimiter{\ceil}{\lceil}{\rceil}

\newcommand{\E}{\mathbb{E}}
\newcommand*\xor{\mathbin{\oplus}}
\usepackage{amssymb,amsmath,amsthm}

\title{Efficient candidate screening
under multiple tests\\ and implications for fairness}

\begin{document}

\author{
  Lee Cohen\thanks{Tel Aviv University. This work was supported in part by The Yandex Initiative for Machine Learning.\\ Email: \href{mailto:leecohencs@gmail.com}{\nolinkurl{leecohencs@gmail.com}}.}\hspace*{1cm}
  Zachary C. Lipton\thanks{Carnegie Mellon University and Amazon AI. This work was supported by the AI Ethics and Governance Fund. Email: \href{mailto:zlipton@cmu.edu}{\nolinkurl{zlipton@cmu.edu}}.}\hspace*{1cm}
  Yishay Mansour\thanks{Tel Aviv University and Google Research. This work was supported in part by a grant from ISF.\\ Email: \href{mailto:mansour.yishay@gmail.com}{\nolinkurl{mansour.yishay@gmail.com}}.}
}

\maketitle

\begin{abstract}
When recruiting job candidates, 
employers rarely observe their underlying skill level directly.
Instead, they must administer a series of interviews
and/or collate other noisy signals
in order to estimate the worker's skill.
Traditional economics papers address screening models
where employers access worker skill via a single noisy signal.
In this paper, we extend this theoretical analysis
to a multi-test setting, considering both Bernoulli and Gaussian models. 
We analyze the optimal employer policy 
both when the employer sets a fixed number of tests per candidate
and when the employer can set a dynamic policy, 
assigning further tests adaptively
based on results from the previous tests.
To start, we characterize the optimal policy 
when employees constitute a single group, 
demonstrating some interesting trade-offs.
Subsequently, we address the multi-group setting, 
demonstrating that when the noise levels vary across groups, 
a fundamental impossibility emerges 
whereby we cannot administer the same number of tests,
subject candidates to the same decision rule,
and yet realize the same outcomes in both groups.
\end{abstract}

\section{Introduction}
\label{sec:intro}
Consider an employer seeking to hire new employees. 
Clearly, the employer would like to hire the best employees for the task, 
but how will she know which are best fit? 
Typically, the employee will gather information on each candidate,
including their education, work history, reference letters, 
and for many jobs, they will actively conduct interviews. 
Altogether, this information can be viewed as the \emph{signal} available to the employer.

Suppose that candidates can be either \emph{skilled} or \emph{unskilled}. 
If the firm hires an ``unskilled'' candidate, 
it will incur a significant cost on account of lost productivity.
For this reason, the employer would like 
to minimize the number of \emph{False Positive} mistakes, 
instances where \emph{unskilled} candidates are hired. 
On the other hand, the employer desires 
not to \emph{overspend} on the hiring process, 
limiting the number of interviews per hired candidate 
(either on average, or absolutely).
However, fewer interviews weakens the signal, 
causing the employer to make more mistakes.
At the heart of our model is this inherent trade-off 
between the quality of the signal 
and the cost of obtaining the signal.
This marks a departure from the classical economics literature,
in which the signal is commonly regarded as a given.

Complicating matters, hiring efficiency is not the only desiderata at play. 
In society, employees belong to various \emph{demographic groups}, 
and we may strive to design policies 
that are in some sense \emph{fair} vis-a-vis group membership. 
While \emph{fairness} can be an elusive notion, 
regulators must translate it to concrete rules and laws.
In the United States, a body of anti-discrimination law
dating to the Civil Rights act of 1964, 
subjects decisions that result in disparate outcomes 
(as delineated by race, age, gender, religion, etc)
to extra scrutiny: employers must not only show 
that preference for some groups over others 
did not drive the decision (disparate treatment doctrine) 
but also justify that any observed disparities 
arise from a business necessity (disparate impact doctrine), 
whether or not those disparities were intentional.

In this paper, we seek to understand how a complex hiring process 
would interact with the requirements of fairness.  
We extend the theory on candidate screening and statistical discrimination,
addressing the setting in which employers can subject 
employees to multiple tests,
which we assume to be conditionally independent
given the worker's skill level and group membership.
To build intuition, most of our analysis focuses 
on a Bernoulli model of both worker skill and screening.
Additionally, we begin to extend the traditionally-studied 
Gaussian skill and screening models to the multi-test setting (Section \ref{sec:gaussian}).

Unlike the classical papers, 
in which an employer's hiring policy is given 
by a simple thresholding rule,
our setting requires greater care 
to derive the optimal employer policy. 
In our setting, we imagine that the employer 
wishes to minimize the number of tests performed
subject to a constraint upper-bounding the false positive rate.
We characterize the optimal policy in this case as a randomized threshold policy.

We also consider the setting in which employers can allocate tests dynamically, 
deciding after each result whether to 
(i) hire the candidate;
(ii) reject the candidate and move on to the next one; or 
(iii) administer a subsequent test. 
In the Bernoulli case, the optimal policy consists 
of administering tests until each candidate's posterior likelihood 
of being a high-skilled worker either 
dips below the prior or rises above a threshold 
determined by the tolerable false positive rate.
We demonstrate that the analysis of this process
can be reduced to a random walk over the log posterior odds
and derive the solution via the corresponding Gambler's ruin problem.

Finally, 
we consider the ramifications for fairness within our model 
when employees, despite possessing similarly-distributed skills, 
are evaluated with differing noise levels.

\subsection{Related work}

The classical economics literature 
on discrimination in employment 
can broadly be divided into two focuses.
The \emph{taste-based discrimination} model 
due to \cite{becker1957economics}
models the market outcomes in a setting 
where employers express an explicit preference 
for hiring members of one group,
acting as if an employee's demographic membership provides utility.
This preference for certain groups
induces a sorting of employees from the disadvantaged group
towards those employers who discriminate the least
with wages ultimately determined by the \emph{marginal discriminator}.
Subsequently, \cite{phelps1972statistical} suggested
a statistical mechanism by which similarly-skilled employees 
from different groups might experience differential outcomes: 
the comparative difficulty of screening from one group vs. another.
Many subsequent works extend this analysis,
typically focusing on Gaussian models of worker quality
and conditionally-Gaussian test scores 
\cite{arrow1973theory, aigner1977statistical}.
These papers consider the setting 
where workers are assessed via a single test
characterized by a group-dependent noise level.
Our work is differentiated from these 
by considering richer mechanisms for acquiring signal.

In the more recent literature on fairness in machine learning,
researchers often focus on binary classification, 
with employees characterized by a protected characteristic (group membership),
and other (non-protected) covariates \citep{pedreshi2008discrimination, kamiran2010discrimination, kamishima2011fairness}.
There, the predictor is presumably used 
to guide a consequential decision,
such as allocating some economic good (loans, jobs, etc.) \citep{corbett2018measure}
or assessing some penalty (e.g. risk scores to guide bail decisions) \citep{chouldechova2017fair}.
Papers then focus on various interventions 
for ensuring accurate prediction subject to various constraints 
such as demographic parity (outcomes independent of group membership), 
blindness (model cannot observe group membership), 
and equalized false negative and/or false positive rates \cite{hardt2016equality}. 
Several simple impossibility results 
preclude simultaneously satisfying 
several combinations of these parities 
\cite{berk2017fairness, chouldechova2017fair, kleinberg2016inherent}.
More recently, a number of papers 
have drawn inspiration from economic modeling, 
extending the literature on fairness in classification 
to consider longer-term dynamics, equilibria, 
and the emergence of feedback loops 
\cite{hu2018short, hardt2016equality, ensign2017runaway}. 
Finally, \cite{Verma2018} provide a survey of definitions 
from the algorithmic fairness literature.

\section{The Bernoulli Model}
\label{sec:bernoulliModel}
We formalize our problem as follows. 
An employer accesses an infinite pool of candidates 
(indexed by $i \in \mathbb{N}^+$), 
each of which has some (hidden) \textit{skill level} $y_i\in\{0,+1\}$,
which denote \emph{unskilled} and \emph{skilled}, respectively. 
Underlying worker skill levels $y_i$ are sampled independently
from a Bernoulli distribution with parameter $p$.
An employer can access information about the $i$-th candidate
through a sequence of $\tau$ tests, 
which are conditionally independent given $y_i$. 
Each \textit{test result}, $\hat{y}_{i,j}\in \{0,+1\}$ 
disagrees with the ground truth skill with probability
$\Pr[\hat{y}_{i,j} \neq y_i]=\frac{1-\sigma}{2}$, 
where $\sigma\in(0,1)$, i.e., $\hat{y}_{i,j}=y_i\xor Br(\frac{1-\sigma}{2})$\footnote{$\xor$ is the XOR operation between two binary random variables, and therefore $\hat{y}_{i,j}$ is also a random variable.}.
For convenience, we denote the noise level as $\eta=\frac{1-\sigma}{2}\in (0,\frac{1}{2})$.
We say that a test result $\hat{y}_{i,j}$ is \textit{flipped} 
if $\hat{y}_{i,j} \ne y_i$, 
and the number of flipped results for a given candidate 
is denoted by 
$Z_\tau^\eta$ is $Z_\tau^\eta=\sum_{j=1}^\tau \mathbb{I}(\hat{y}_{i,j}\ne y_i)$, 
where $\mathbb{I}(\cdot)$ is the indicator function.

A \emph{selection criterion} is a mapping 
between test results to actions: $\text{\tt Select}(\hat{y}_{i,1},\dots,\hat{y}_{i,\tau_i}) \in \{0,1\}$, where $0$ means \emph{reject} 
and $1$ means \emph{accept} (hire). 
A \emph{policy} $\pi$ sets the selection criteria 
based on $\sigma, p$ and other possible constraints 
such as probability to hire, error probability, etc. 
A \emph{randomized threshold policy} is a policy $\pi$ with parameters $(\tau,\theta,r)$ such that $\pi(\hat{y}_{i,1},\dots,\hat{y}_{i,\tau_i})=1$ 
for $S_{\tau}:=\sum_{i=1}^{\tau} \hat{y}_{i,j} > \theta$,
$\pi(\hat{y}_{i,1},\dots,\hat{y}_{i,\tau_i})=0$ 
for $S_{\tau}< \theta$, and for $S_{\tau}= \theta$ the probability that $\pi(\hat{y}_{i,1},\dots,\hat{y}_{i,\tau_i})=1$ is $r$.
We call a policy $\pi$ a \emph{threshold policy} if $r=1$.
In a \textit{dynamic policy}, 
rather than setting a fixed number of tests per candidate,
the employer may decide after each test whether to 
\emph{accept}, \emph{reject}, 
or to perform an additional test, i.e., $\pi(\hat{y}_{i,1},\dots,\hat{y}_{i,\tau_i}) \in \{0,1,\texttt{more}\}$. 
Note that for a dynamic policy, the number of tests $\tau$
is a random variable determined based on the tests' outcomes.
When designing a policy, one must carefully consider 
the balance between the following desiderata:
\begin{enumerate}
\item \textbf{Minimize False Discovery Rate (FDR)}---the fraction of unskilled workers 
among the accepted candidates, i.e., $\text{FDR}:=\Pr[y_i=0|\pi(\hat{y}_{i,1},\dots,\hat{y}_{i,\tau})=+1]$.
\item \textbf{Minimize False Omission Rate (FOR)}---the fraction of skilled workers among the rejected candidates, i.e., $\text{FOR}:=\Pr[y_i=+1|\pi(\hat{y}_{i,1},\dots,\hat{y}_{i,\tau})=0]$. 
\item \textbf{Minimize False Negatives (FN)}---the amount of skilled workers that are classified as unskilled.
\item \textbf{Minimize False Positives (FP)}---the amount of unskilled workers that are classified as skilled.
\item \textbf{Ratio of accept probability and the number of tests}---the number 
of tests performed per candidate hired, 
using a parameter $B>1$, we have  $\frac{\tau}{B} \leq \Pr[\pi(\hat{y}_{i,1},\dots,\hat{y}_{i,\tau})=+1]$.
\end{enumerate}
For any fixed 
number of tests $\tau$, 
increasing the threshold $\theta$ of a threshold policy
decreases FDR while increasing FOR.

\noindent{\bf Loss:} 
To handle the trade-off between the false positives, 
(i.e., when an unskilled candidate is accepted) 
and false negatives 
(i.e., when a skilled candidate is rejected), 
we introduce an $\alpha$-loss, 
paramaterized by $\alpha \in [0,1]$ and defined as follows:
\begin{equation*}
\ell_{\alpha}(b_1,b_2)=\alpha \mathbb{I}[b_1=0,b_2=1] + (1-\alpha)\mathbb{I}[b_1=1,b_2=0]
\end{equation*}
where $\mathbb{I}[\cdot]$ is the indicator function and $b_1,b_2\in\{0,1\}$. The expected loss of a policy $\pi$ is,
\begin{equation}\label{loss}
    l_\alpha(\pi)=E[\ell_{\alpha}(y_i,\pi(\hat{y}_{i,1},\dots,\hat{y}_{i,\tau}))] 
\end{equation}
where the expectation is over the type of the candidates $y_i$, the test results $\hat{y}_{i,j}$, and the decisions of $\pi$.

\section{Analysis of the Bernoulli Model with One Group}
\label{sec:oneGroup}
To begin, we analyze this hiring model 
for a single group of candidates.
The employer's goal is to minimize the expected loss, $l_{\alpha}(\pi)$, 
while maintaining a given acceptance probability.
For brevity, we relegate all proofs to the appendix.

\subsection{The Simple Threshold Policy (Equal Number of Tests)}

Consider the setting where 
the employer must subject all candidates 
to an equal number of tests $\tau$ and threshold $\theta$ 
(these parameters are chosen by the employer but thereafter constant across candidates).
For a given threshold, we can relate
the flip probability (error rate) of the test
to the probability that a candidate is accepted as follows:

Recall that $\hat{y}_{i,j}=y_i\xor Br(\eta)$, $S_\tau=\sum_{j=1}^\tau \hat{y}_{i,j}$, that $Z^\eta_\tau=\sum_{t=1}^\tau \mathbb{I}(\hat{y}_{i,j}\neq y_i)$, and that $\tau$ and $\theta$ are the only parameters of the threshold policy, $\pi$.
Informally, $S_\tau$ is the number of passed tests
and $Z^\eta_\tau$ is the number of flips 
(tests in error).
The probability of hiring an unskilled candidate is given by:
$$\Pr[\pi(\hat{y}_{i,1},\dots,\hat{y}_{i,\tau})=1|y_i=0]=\Pr[S_{\tau}\geq \theta|y_i=0]=
\Pr[Z_{\tau}^{\eta}\geq \theta].$$

Since $Z_{\tau}^{\eta}$ is a binomial random variable with parameters $\tau$ and $\eta$, we can calculate this probability precisely as:
$\Pr[\pi(\hat{y}_{i,1},\dots,\hat{y}_{i,\tau})=1|y_i=0]=\Pr[Z_{\tau}^{\eta}\geq \theta]= \sum_{k=\theta}^\tau{\tau \choose k}\eta^k(1-\eta)^{\tau-k}$,
and the probability of rejecting a skilled candidate is the probability that they encounter more than $\tau-\theta$ flips, thus: 
$\Pr[\pi(\hat{y}_{i,1},\dots,\hat{y}_{i,\tau})=0|y_i=+1]=\Pr[S_{\tau}<\theta|y=+1]=\Pr[Z_\tau^{\eta}> \tau- \theta]=$\\
$\sum_{k=\tau- \theta+1}^{\tau}{\tau \choose k}\eta^k(1-\eta)^{\tau-k}.$
Similarly, given a candidate's skill level, 
we can calculate the probability 
that 
they obtain exactly $k$ positive tests 
out of $\tau$, i.e,
$$\Pr[S_{\tau}=k|y_i=0]=\Pr[Z_\tau^{\eta}= k]={\tau \choose k}\eta^{k}(1-\eta)^{\tau-k}.$$
$$\Pr[S_{\tau}=k|y_i=+1]=\Pr[Z_\tau^{\eta}= \tau- k]={\tau \choose k}\eta^{\tau-k}(1-\eta)^{k}.$$
Given these observations, we can now analyze the employer's choices. 

\subsection*{Optimal solution for any ratio $\alpha\in(0,1)$}
The next theorem shows that for threshold policies, the  expected loss $l_{\alpha}(\pi)=l_{\alpha}(\theta)$ is minimized at $\theta^*_{p,\alpha}$ 
such that
$|\theta^*_{p,\alpha}-\tau/2|\leq \frac{\log (\frac{1}{p})+\log(\frac{1}{ \alpha})}{2\log(1+\frac{2\sigma}{1-\sigma})}$.
\begin{theorem}\label{thm:minErrorThresholdgeneralAlpha}
The loss function $l_{\alpha}(\theta)$ is quasi-convex and a
threshold of 
$\theta^*_{p,\alpha}=
\left\lceil\frac{1}{2}\left(\tau-\frac{\log(\frac{1}{p}-1)+\log(\frac{1}{\alpha}-1)}{\log(1+\frac{2\sigma}{1-\sigma})}\right)\right\rceil$ 
minimizes loss for any values of $\alpha,p,\sigma\in(0,1)$.
Namely,
\[
\theta^*_{p,\alpha}=\arg\min_{\theta}l_{\alpha}(\theta) =\left\lceil\frac{\tau}{2}-\frac{\log(\frac{1}{p}-1)+\log(\frac{1}{\alpha}-1)}{2\log(1+\frac{2\sigma}{1-\sigma})}\right\rceil.
\]
\end{theorem}

Next, we bound the number of tests required to guarantee that
the probability of classification error by the majority decision rule (i.e., $\theta=\ceil{\frac{\tau}{2}}$) 
does not exceed a specified quantity $\delta$.

\begin{theorem}\label{thm:majorityLowerBound}
For every $\delta,p,\alpha \in(0,1)$, performing 
$\tau=\Omega(\frac{\alpha+p-2p\alpha}{\sigma^2}\ln(\frac{1}{\delta}))$ 
tests per candidate 
and using majority as a decision rule (i.e., $\theta=\tau/2$)
guarantees $l_{\alpha}(\pi)\leq \delta$.
\end{theorem}

\subsubsection*{Equal cost for false positives and false negatives $(\alpha=\frac{1}{2})$}

Consider the simple loss consisting of the classification error rate 
(false positives and false negatives count equally), 
expressed via our loss function by setting $\alpha=\frac{1}{2}$. 
When skilled and unskilled candidates 
occur with equal frequency, i.e., $p=1/2$, 
we can derive that the majority decision rule 
minimizes the classification error 
for any number of tests.

\begin{corollary}\label{cor:majorityMinError}
Assume  $p=1/2$ and $\alpha=1/2$. 
For any number of tests $\tau$, 
the majority decision rule minimizes loss $l_\alpha$.
Namely,
$\arg\min_{\theta}l_{\frac{1}{2}}(\theta) =\ceil{\frac{1}{2}\tau}.$
In addition, for every $\delta\in(0,1)$, 
performing $\tau=\Omega(\frac{1}{\sigma^2}\ln(\frac{1}{\delta}))$ tests per candidate 
and using majority as a decision rule 
guarantees classification error with probability of at most $\delta$.
\end{corollary}

\paragraph*{FDR minimization with limited number of tests per hire for balanced groups}
Again, assuming balanced groups (i.e., $p=1/2$),
suppose that an employer would like 
to minimize the false discovery rate, 
subject to the constraint of lower bounding the hiring probability.
We can model this optimization problem 
by introducing a budget parameter $B>1$ 
to bound any predetermined (fixed) number of tests 
per hired candidate as follows:
\begin{equation}\label{optimization1}
\begin{aligned}
& \arg\min_{\pi}
& & \text{FDR}_{\pi}=\Pr[y_i=0|\Pr[\pi(\hat{y}_{i,1},\dots,\hat{y}_{i,\tau}) =1] \\
& \text{subject to}
& & \frac{\tau_\pi}{ \Pr[\pi(\hat{y}_{i,1},\dots,\hat{y}_{i,\tau}) =1]}\leq B
\end{aligned}
\end{equation}
where $\tau_\pi$ is the number of tests $\pi$ performs.
The following theorem shows that the optimal policy is a randomized threshold policy.
\begin{theorem}\label{thm:minFPwithBudget}
There exists a randomized threshold policy $\pi$ 
which is an optimal solution for (\ref{optimization1}).
\end{theorem}
\subsection{The Dynamic Policy (Adaptively-Allocated Tests)}
Recall that under a dynamic policy, 
the employer can decide after each test 
whether to accept, reject, or perform another test.
In general, dynamic policies are more efficient 
than those that must set a fixed number of tests. 
To build intuition,
consider a candidate that has passed $2$ out of $3$ tests.
As seen above, under an optimally-constructed fixed-test policy,
any candidate that fails a single test might be rejected.
\footnote{For example, if $B=18$ and $\eta=\frac{1}{3}$,
the lowest false discovery rate is achieved by $\tau=\theta=3$.}
However, the posterior probability 
that this candidate is in fact \emph{skilled}
may still be greater than that of a fresh candidate sampled from the pool.
Thus we can improve on the fixed-test policy by dynamically
allocating more tests to candidates 
until their posterior odds either 
dip below the prior odds or rise above the threshold for hiring.
The following theorem formalizes this notion
that it is better to administer more tests 
to a candidate that passed the majority of previous tests 
than to start afresh with a new candidate:

\begin{theorem}\label{thm:posteriorBetter}
For any $p,\sigma, \tau$, a candidate $i$ 
that passed $\theta>\frac{\tau}{2}$ out of $\tau$ tests 
is more likely to be a skilled
than a freshly-sampled candidate $i'$ 
for whom no test results are yet available, i.e.,
$\Pr[y_{i'}=+1]=p<\Pr[y_i=+1|S_{\tau}=\theta]$.
\end{theorem}
\begin{remark}
If $\theta<\frac{\tau}{2}$, the inequality would have been reversed.
\end{remark}

\paragraph*{The Greedy Policy}
We now present a greedy algorithm 
that continues to test a candidate 
so long as the posterior probability that $y_i=+1$ 
is greater than $\epsilon'$ and smaller than $1-\epsilon$,
rejects a candidate whenever the posterior falls below $\epsilon'$
(absent fairness concerns, employers will set $\epsilon'=p$ for all groups), 
and accepts whenever the posterior rises above $1-\epsilon$. 
Given parameters $\epsilon,\epsilon'>0$, 
we show that the greedy policy solves the optimization problem 
of minimizing the mean number of tests under these constraints, i.e.,
\begin{equation*}
\begin{aligned}
& \underset{\tau}{\text{minimize}}
& & \E[\tau] \\
& \text{subject to}
& &\forall_i \pi(\hat{y}_{i,1},\dots,\hat{y}_{i,\tau}) =1 \text{ iff } \Pr[y_i=+1|\hat{y}_{i,1},\dots,\hat{y}_{i,\tau}]\geq 1-\epsilon\\
& & & \forall_i \pi(\hat{y}_{i,1},\dots,\hat{y}_{i,\tau}) =0 \text{ iff } \Pr[y_i=+1|\hat{y}_{i,1},\dots,\hat{y}_{i,\tau}]<\epsilon'
\end{aligned}
\end{equation*}
Our analysis of this policy builds upon the observation
that conditioned on a worker's skill,  
the posterior log-odds after each test
perform a one-dimensional random walk,
starting with the prior log-odds $\log(\frac{p}{1-p})$
and moving, after each test result, either left 
(upon a failed test) or right (upon a passed test).  
When (as in our model) the probability of a flip 
are equal for skilled and unskilled candidates, 
our random walk has a fixed step size.
Moreover, our random walk has \emph{absorbing barriers}
corresponding to (when $\epsilon'=p$) 
falling below the prior log odds (on the left)
and exceeding the hiring threshold (on the right).
Owing to the fixed step size and absorbing barriers,
our policy resembles the classic problem of Gambler's ruin,
in which a gambler wins or loses a unit of currency at each step,
and loses when crossing a threshold on the left (going bankrupt)
or on the right (bankrupting the opponent).
We formalize the random walk as follows where $X_j$
is the position on the walk at time $j$:
\begin{enumerate}
    \item $X_0$ is the prior log-odds of the candidate, i.e., $X_0=\log \frac{p}{1-p}$.
    \item After each test result, $\hat{y}_{i,j}$ is observed,
    $X_j=X_{j-1}+(2\hat{y}_{i,j}-1)\cdot
    \log \left(\frac{\Pr[\hat{y}_{i,j}=+1|y_i=+1]}{\Pr[\hat{y}_{i,j}=+1|y_i=0]}\right)$.
\end{enumerate}
Let $\pi_{Greedy}$ be the policy that accepts a candidate if $\Pr[y_i=+1|\hat{y}_{i,1},\dots,\hat{y}_{i,j}]\geq 1-\epsilon$, 
rejects if $\Pr[y_i=+1|\hat{y}_{i,1},\dots,\hat{y}_{i,j}]<\epsilon'$,
and otherwise conducts an additional test, i.e.,
\[
\pi_{Greedy}(\hat{y}_{i,1},\dots,\hat{y}_{i,j})=
 \begin{cases}
       0    &\quad\text{if } \Pr[y_i=+1|\hat{y}_{i,1},\dots,\hat{y}_{i,j}]<\epsilon' \\
       1    &\quad\text{if } \Pr[y_i=+1|\hat{y}_{i,1},\dots,\hat{y}_{i,j}]\geq 1-\epsilon\\
       \text{retest}    &\quad\text{else}\\
     \end{cases}.
\]
An employer will generally set
the lower absorbing barrier
to reject all candidates with posterior log odds less than $p$ 
since a fresh candidate from the pool is expected to be better.
However, when noise levels differ across groups,
we may prefer \emph{in the interest of fairness} 
to set $\epsilon'$ lower than $p$ for members of the noisier group, 
allowing us to equalize the frequency of false negatives across groups (see Section \ref{sec:twoGroups}).

\begin{lemma}\label{lemma:beta}
Let $\beta,\beta'\in \mathbb{R}$ be the parameters that satisfy $\frac{\beta}{\beta+1}=1-\epsilon$ and $\frac{\beta'}{\beta'+1}=\epsilon'$
(i.e., $\beta=\frac{1-\epsilon}{\epsilon}$and $\beta'=\frac{\epsilon'}{1-\epsilon'}$).
Then $X_{\tau}\geq \log \beta$ 
iff $\Pr[y_i=+1|\hat{y}_{i,1},\dots,\hat{y}_{i,\tau} ] \geq 1-\epsilon$ 
(iff the candidate is accepted) and $X_{\tau}< \log \beta'$ 
iff $\Pr[y_i=+1|\hat{y}_{i,1},\dots,\hat{y}_{i,\tau} ] < \epsilon'$ 
(iff the candidate is rejected).
\end{lemma}

\begin{corollary}\label{cor:piGreedy}
The policy $\pi_{Greedy}$ can be described as follows.
\[
\pi_{Greedy}(\hat{y}_{i,1},\dots,\hat{y}_{i,\tau})=
 \begin{cases}
       0     &\quad\text{if } X_{\tau} < \log \frac{\epsilon'}{1-\epsilon'} \\
       1    &\quad\text{if } X_{\tau}\geq \log(\frac{1-\epsilon}{\epsilon})\\
       \textnormal{retest}    &\quad\textnormal{else}\\
     \end{cases}
\]
\end{corollary} 

\noindent We use the following parameters in the next theorems: 
$$a=\left\lceil \frac{\log(\frac{(1-\epsilon)(1-\epsilon')(1+\sigma)}{\epsilon \epsilon'(1-\sigma)})}{\log(\frac{1+\sigma}{1-\sigma})}\right\rceil \gg \frac{1}{\sigma}
\text{\quad and \quad} z=
\left\lceil\frac{\log (\frac{p(1-\epsilon')(1+\sigma)}{\epsilon'(1-p)(1-\sigma)})}{\log(\frac{1+\sigma}{1-\sigma})}\right\rceil$$
\begin{theorem}[Expected number of tests per type]\label{thm:expectedNumOfTests}
The expected number of tests until a decision 
(namely accept or reject) for skilled candidates is $\E[\tau_{s}]=\frac{1}{\sigma}\left(a\cdot\frac{1-(\frac{1-\sigma}{1+\sigma})^{z}}{1-(\frac{1-\sigma}{1+\sigma})^{a}}-z\right)\approx\frac{2a}{1+\sigma}-\frac{z}{\sigma}$ and $\E[\tau_{u}]=\frac{1}{\sigma}\left(z-a\cdot\frac{1-(\frac{1+\sigma}{1-\sigma})^{z}}{1-(\frac{1+\sigma}{1-\sigma})^{a}}\right)\approx\frac{z}{\sigma}$ for unskilled candidates.
\end{theorem} 
For the probabilities of the candidates to be accepted or rejected, 
conditioned on their true skill level, 
we present the results in a form of confusion matrix in Table \ref{tab:confMat}.

\begin{table}
\begin{center}
\caption{Confusion matrix for $\pi_{\text{greedy}}$
assuming $\epsilon\leq 1/4$ and $\epsilon'\leq p\leq 1/2$.
}
\small
\begin{tabular}{l|ll|ll}
\multicolumn{1}{c}{} &\multicolumn{2}{c}{\textbf{General $\epsilon'$}}&\multicolumn{2}{c}{\textbf{When $\epsilon' = p$}} \\ 
\multicolumn{1}{c}{} & 
\multicolumn{1}{c}{\textbf{Skilled}  ($y_i=+1$)} & 
\multicolumn{1}{c}{\textbf{Unskilled} ($y_i=0$)} &
\multicolumn{1}{c}{\textbf{Skilled} } & 
\multicolumn{1}{c}{\textbf{Unskilled} }
\\ 
\toprule
\textbf{accept} & $\text{TPR}=\Theta\left(1-\frac{\epsilon'}{p}(1-\sigma)\right)$ & $\text{FPR}=\Theta\left(\epsilon(p-\epsilon'+\epsilon'\sigma) \right)$  &$ \Theta(\sigma)$&$ \Theta(\epsilon p \sigma)$ \\
\textbf{reject} & $\text{FNR}=\Theta\left(\frac{\epsilon'}{p}(1-\sigma)\right)$   & $\text{TNR}=\Theta\left(1-\epsilon(p-\epsilon'+\epsilon'\sigma) \right)$ & $ \Theta(1-\sigma)$ &$\Theta(1-\epsilon p \sigma)$\\ 
\bottomrule
\end{tabular}
\label{tab:confMat}
\end{center}
\end{table}

\begin{theorem}\label{thm:expectedNumOfTestsUntilDecision}
The expected number of tests until deciding whether to accept or reject a candidate is $
\E[\tau|{\pi(y_{i,\tau})\in \{0,1\}}]
\approx \frac{ap}{\sigma}$, where 
$a\gg\frac{1}{\sigma}$.
\end{theorem}

\section{Fairness Considerations in the Two-Group Setting}
\label{sec:twoGroups}
\paragraph{Two Groups---Threshold Policies}

We now discuss the effects of a threshold policy 
when candidates belong to two groups, $G_1$ and $G_2$ 
whose skill level is distributed identically,
but whose tests are characterized by different noise levels.
Without loss of generality, we assume that $\eta_1<\eta_2$, 
where $\eta_i$ is the probability 
that a test result of a candidate from $G_i$ is different from his skill level.
To begin, we note the fundamental irreconcilability of equalizing either the false positive or the false negative rates across groups with subjecting candidates to the same policy.

\begin{theorem}[Impossibility result]\label{thm:stochasticDominance}
When noise levels differ between two groups with identical skill level distribution, 
a single Threshold Policy $\pi$ 
(with the same number of tests $\tau$ 
and the same threshold $\theta$ for both groups) 
cannot have equality in either the false negative rates 
or in the false positive rates across the groups. 
Particularly, there is a higher false positive rate in the noisier group, 
as an unskilled candidate from $G_2$ is more likely to be accepted 
by the threshold policy than an unskilled candidate from $G_1$:
\begin{equation*}
\textnormal{FPR}_{\theta,\tau}^{\eta_1}=
\Pr_{\eta_1}[\pi(\hat{y}_{i,1},\dots,\hat{y}_{i,\tau})=1|y_i=0]<\Pr_{\eta_2}[\pi(\hat{y}_{i,1},\dots,\hat{y}_{i,\tau})=1|y_i=0]
=\textnormal{FPR}_{\theta,\tau}^{\eta_2},
\end{equation*}
and also a higher false negative rate, 
as a skilled candidate from $G_2$ is more likely to be rejected 
than a skilled candidate from $G_1$:
\begin{equation*}
\textnormal{FNR}_{\theta,\tau}^{\eta_1}=
\Pr_{\eta_1}[\pi(\hat{y}_{i,1},\dots,\hat{y}_{i,\tau})=0|y_i=+1]<\Pr_{\eta_2}[\pi(\hat{y}_{i,1},\dots,\hat{y}_{i,\tau})=0|y_i=+1]=\text{FNR}_{\theta,\tau}^{\eta_2}.
\end{equation*}
\end{theorem}
\textbf{Connection to Economics Literature }%
 Aigner and Cain \cite{aigner1977statistical} discuss 
a similar case under a Gaussian screening model
where the variance (noise level) of the single test
differs across the two groups. 
Similarly, they note that qualified candidates fare worse in the noisy group
but that unqualified candidates fare better in the noisier group.
Our work differs from theirs in that 
we consider the effect of multiple tests
and the ability to optimize over the number of tests.

\paragraph{Two Groups--Dynamic policy}
We now consider the (dynamic) hiring policy 
in the setting when employees belong to two groups, 
$G_1$ and $G_2$ with identically-distributed skills 
but different noise levels $\eta_1<\eta_2$.
We note that there are two ingredients 
that explain the differences among the groups:
(i) The step size, $\log \left(\frac{\Pr[\hat{y}_{i,j}=+1|y_i=+1]}{\Pr[\hat{y}_{i,j}=+1|y_i=0]}\right)=\log \left(\frac{1-\eta}{\eta} \right)$ of $G_2$ (the noisier group) 
is smaller than the step size of $G_1$. 
Thus these candidates must typically pass more tests 
before they are accepted;
and (ii)
Skilled candidates in group $G_2$ 
exhibit less drift to the right
(they have a higher probability of failing a test).
Consequently, when an employer (rationally)
sets $\epsilon' = p$ for all groups,
a skilled candidate from $G_2$ 
is more likely to be fail a test in step $1$,
at which point the dynamic policy summarily rejects them.
These two facts explain both the higher false negative rates for $G_2$ 
and the longer expected duration until acceptance.
By setting $\epsilon' < p$ for members of the noisier group, we can equalize false negative rates.
Precisely, setting $\epsilon' = \frac{\eta_1}{\eta_2}p$
achieves the desired parity.
The cost of this intervention is that it requires more tests for candidates from the noisier group.
Here, our random walk analysis can be leveraged to determine exactly how many more.
Once again, we cannot provide equality across the groups in all desired ways---the 
same acceptance criterion, the same expected number of tests,
and the same false negative rates between groups---with the noise differs across groups.

\section{Gaussian Worker Screening Model}
\label{sec:gaussian}
In this section, we work out the analytic solutions for the conditional expectation of worker qualities given a series of conditionally independent tests $Y_1, ... Y_n$ s.t. $\forall i,j$, $Y_i \perp  Y_j |Q$.
We assume that the worker quality $Q$ normally distributed with mean $\mu_Q$ and variance $\sigma_Q^2$, so instead of binary skill level we have continuous quality of candidates.
Conditioned on $Q=q$, each test is generated according to the structural equation 
$y_i = q + \eta$, where $\eta$ is a normally distributed noise term
with mean $0$ and variance $\sigma_\eta^2$.
Equivalently, we can say that the conditional 
distribution for each test $P(Y|Q=q)$
is Gaussian with mean $q$ and variance $\sigma_\eta^2$. 
We refer the reader to Appendix \ref{app:gaussianExtended}
for further details.

\label{sec:gaussianTwoGroups}
We show that we can equalize conditional variance  between the two groups by giving more interviews to to noisier group, and that it yields the same conditional expectations.

\begin{theorem}\label{thm:EqVariances}
For two groups, $G_1,G_2$ with the same worker quality $Q$, that differ only in the variance of their noise 
$\sigma_{\eta_1}^2<\sigma_{\eta_2}^2$, the variance can be equalized by using $n_2=\frac{\sigma_{\eta_2}^2}{\sigma_{\eta_1}^2}n_1$ interviews (or tests) for $G_2$, where $n_1$ is the number of interviews for each candidate from $G_1$. 
\end{theorem}

\begin{theorem}\label{thm:EqExpectations}
When equalizing conditional variances between $G_1,G_2$ by using $n_2=\frac{\sigma_{\eta_2}^2}{\sigma_{\eta_1}^2}n_1$, we get the same conditional expectations,  $\mathbbm{E}_{\eta_1}[Q|Y_1,...,Y_{n_1}]=\mathbbm{E}_{\eta_2}[Q|Y_1,...,Y_{n_2}]$.
\end{theorem}

\section{Unsupervised Parameter Estimation}
\label{sec:unsupervised}

Now, under the assumption of realizable case, 
we explain how one can estimate the parameters $p$ and $\sigma$
given tests results from a homogeneous population.
Surprisingly, we discover that parameter recovery
in this model does not require any ground truth labels
indicating whether an employee is skilled or unskilled.
We use Hoeffding’s inequality to bound the absolute difference 
between the estimated parameters and the true parameters 
by choosing $\delta$ as the wanted upper bound 
and solving for the number of samples or $\epsilon$.
\begin{lemma}[Hoeffding’s inequality]
Let $y_1,\dots,y_m$ be $\sigma^2-$sub--gaussian random variables. 
Then, for any $\epsilon>0$,
\[
\Pr\left[\left|\frac{1}{m}\sum_{i=1}^m y_i-\E[y_i]\right|\geq \epsilon\right]\leq 2e^{-m\epsilon^2/2\sigma^2}.
\]
If $y_1,\dots,y_m$ are Bernoulli random variables with parameter $p$,
\[
\Pr\left[\left|\frac{1}{m}\sum_{i=1}^m y_i-p\right|\geq \epsilon\right]\leq 2e^{-2m\epsilon^2}.
\]
\end{lemma}
We start by estimating $\sigma$ and then use it to derive an estimate for $p$. 
The estimated parameters are denoted by $\hat{\sigma}$ and $\hat{p}$.
Notice that in order to have any information regarding the true value of $\sigma$, 
we need to have candidates with at least two tests. 
Hence, from now on we assume exactly that, i.e., 
$\forall_i \pi_{\text{Greedy}}(\hat{y}_{i,1})=more$ for dynamic policies 
and $\tau\geq2$ for fixed number of tests policies. 

Now, in both policies we have showed that the optimal rule 
is to reject candidates that fail their first test. 
Therefore inconsistencies between the first two tests 
are seen only in cases where $\hat{y}_{i,1}=1,\hat{y}_{i,2}=0$.

Let $c$ be the number of inconsistencies in the first two tests, 
i.e., $c=|\{(\hat{y}_{i,1},\hat{y}_{i,2}):y_{i,1}\ne y_{i,2}\}|$, 
and let $m$ be the number of candidates with at least two tests. 
Since $c$ is generated by sampling $m$ times,
the distribution $Br((\frac{1+\sigma}{2})(\frac{1-\sigma}{2}))=Br(\frac{1-\sigma^2}{4})$ and we can estimate $\sigma$ as stated in the next theorem:
\begin{theorem}
If we have results from $m\geq \frac{1}{2\epsilon^2} \ln \frac{2}{\delta}$ candidates, 
by using $\hat{\sigma}=\sqrt{1-4\frac{c}{m}}$,
then with probability $1-\delta$ we have that $|\hat{\sigma}-\sigma|\leq \epsilon$.
\end{theorem}
Having an estimation of the parameter $\hat{\sigma}$, 
we can calculate the estimated $p$ as follows:
Let $p_{\hat{y}_{*,1}=1}:=\frac{\sum_i\mathbb{I}(\hat{y}_{i,1}=1)}{m}$ be 
the percentage of positive first tests. 
Since this number is generated by the distribution $Br(\frac{1}{2}(p(1+\sigma)+(1-p)(1-\sigma)))=Br(\frac{1}{2}+(2p-1)\frac{\sigma}{2})$, 
we can estimate $\hat{p}$ using the estimated value of $\hat{\sigma}$.

\begin{theorem}
If we have results from $m\geq \frac{1}{2\epsilon^2} \ln \frac{2}{\delta}$ candidates, 
by using $\hat{p}=\frac{2(p_{y_{*,1}=1}-1)+\hat{\sigma}}{\hat{{\sigma}}}$,
we get that with probability $1-\delta$ we have that $|\hat{p}-p|\leq 2\epsilon$.
\end{theorem}
Under the Gaussian screening model,
the parameter estimation is also straightforward 
(assuming realizability) without access to the true skill level of the employees. 
We start by looking at a single candidate, $i$. 
Each of his test results, $\hat{y}_{i,j}$ is generated 
from a conditional distribution $P(Y_{i}|Q_i=q_i)$ 
which is a Gaussian with mean $q_i$ and variance $\sigma_\eta^2$. 
Since this variance is common among all the candidates, 
we can simply average the estimated variance of every candidate 
to get an approximation for $\sigma_\eta^2$.
Suppose $\hat{y}_{i,1},\dots,\hat{y}_{i,n}$ is a sequence of $n$ i.i.d tests of candidate $i$, 
and let $\boldsymbol{y_i}=\frac{1}{n}\sum_{j=1}^n y_{i,j}$ 
be the empirical mean of candidate $i$'s tests.

The following theorem is a result from Hoeffding’s Inequality, 
in which we use to bound the error of our estimated parameters.
\begin{theorem}
By using the following as estimators for Gaussian parameters $\hat{\mu}_Q=\frac{1}{m}\sum_{i=1}^m\boldsymbol{y_i}$,
$\hat{\sigma}_{\eta}^2=\frac{1}{m}\sum_{i=1}^m\frac{1}{n}\sum_{j=1}^{n} (y_{i,j}-\boldsymbol{y_i})^2$ and $\hat{\sigma}_{Q}^2=\frac{1}{m}\sum_{i=1}^m (\hat{\mu}_Q-\boldsymbol{y_i})^2$
(notice that $\E[\hat{\sigma}_{\eta}^2]={\sigma}_{\eta}^2$ and  $\E[\hat{\sigma}_{Q}^2]={\sigma}_{Q}^2$), the difference between each parameter and it's estimator is bounded by $O(\sqrt{\frac{1}{m}\ln(\frac{1}{\delta})})$.
\end{theorem}

\section{Discussion and Future Work}
\label{sec:Summary}
Consider two groups with identically-distributed skills
and characterized by different noise levels in screening. 
Our results demonstrate that if a regulatory body
(e.g., policymakers or a regulator) 
insists on the same number of tests 
and the same decision rule for both groups, 
this would yield higher false positive rates in any threshold policy. 
As a result, hired candidates from the noisier group 
would suffer higher rates of firing.
In turn, this might lead employers to erroneously conclude 
that this group's skill level is lower than it actually is.
This paper presents a policy that handles this problem 
by minimizing the false positive rates of both groups, 
in the form of a greedy policy. 
Moreover the greedy policy is efficient,
minimizing the expected number of tests per hire
among all policies that achieve a specified false positive rate and continue testing every candidates that appear better than the a new one.
However, the dynamic policy will still suffer 
(as does the simple threshold policy)
from higher false negative rates for the noisier group,
violating a notion of fairness dubbed 
\emph{equality of opportunity} in the recent literature 
on fairness in machine learning \cite{hardt2016equality}. 
We addressed this problem by modifying the greedy policy 
to reject candidate
iff 
$\Pr[y_i=+1|\hat{y}_{i,1}\dots \hat{y}_{i,\tau}]<\epsilon'$ by setting $\epsilon' < p$.
Our greedy policy can be made  forgiving and equalize false negative rates across groups.
In future work, we plan to explore extensions to the Gaussian model.

\newpage
\bibliographystyle{apa-good}
\bibliography{refs}

\newpage
\appendix

\section{ Proofs}
\label{sec:proofs}

\subsection{Proofs from Section \ref{sec:oneGroup}}
\label{app:oneGroup}
\begin{proof}[Proof of Theorem \ref{thm:minErrorThresholdgeneralAlpha}]
To prove the theorem, we show that the loss function $l_{\alpha}(\tau,\theta)$, as a function of $\theta$  is quasi-convex and achieves its minimum value at $\left\lceil \frac{1}{2}(\tau-\frac{\log(\frac{1}{p}-1)+\log(\frac{1}{\alpha}-1)}{\log(1+\frac{2\sigma}{1-\sigma})})\right\rceil$.\\
Namely, we show that the loss is monotone increasing for  $\left\lceil \frac{1}{2}(\tau-\frac{\log(\frac{1}{p}-1)+\log(\frac{1}{\alpha}-1)}{\log(1+\frac{2\sigma}{1-\sigma})})\right\rceil\leq \theta \leq \tau-1$, i.e.,
increasing $\theta$ increases the loss: $l_{\alpha}(\theta)<l_{\alpha}(\theta+1)$.\\
Similarly, we show that for $1\leq  \theta\leq \left\lceil \frac{1}{2}(\tau-\frac{\log(\frac{1}{p}-1)+\log(\frac{1}{\alpha}-1)}{\log(1+\frac{2\sigma}{1-\sigma})}) \right\rceil$, we have
$l_{\alpha}(\theta)<l_{\alpha}(\theta-1)$.\\
Indeed,
\[
l_{\alpha}(\theta+1,\tau)-l_{\alpha}(\theta,\tau)=
-\alpha\Pr[y=0,S_{\tau}=\theta] +(1-\alpha)\Pr[y=+1,S_{\tau}=\theta] \]
\[
=-\alpha \Pr[S_{\tau}=\theta|y=0]
\Pr[y=0]+
(1-\alpha)\Pr[S_{\tau}=\theta|y=+1]
\Pr[y=+1]\]
Since $\Pr[y=0]=1-p$ and $\Pr[y=+1]=p$,  we have
\[
l_{\frac{1}{2}}(\theta+1,\tau)-l_{\frac{1}{2}}(\theta,\tau)=-(1-p)\alpha\Pr[S_{\tau}=\theta|y=0]+
p(1-\alpha)\Pr[S_{\tau}=\theta|y=+1].
\]
The above expression is positive iff
\begin{equation}\label{positiveiffgeneral}
    (1-p)\alpha\Pr[S_{\tau}=\theta|y=0]<
p(1-\alpha)\Pr[S_{\tau}=\theta|y=+1]
\end{equation}
Since $\Pr[S_{\tau}=\theta|y=0]$ is the probability of exactly $\theta$ flips, and $\Pr[S_{\tau}=\theta|y=+1]$ is the probability of exactly $\tau-\theta$ flips, we can calculate those probabilities as follows: 
\[
\Pr[S_{\tau}=\theta|y=0]={\tau \choose \theta}(\frac{1-\sigma}{2})^\theta(\frac{1+\sigma}{2})^{\tau-\theta}
\]
\[
\Pr[S_{\tau}=\theta|y=+1]={\tau \choose \tau-\theta}(\frac{1-\sigma}{2})^{\tau-\theta}(\frac{1+\sigma}{2})^{\theta}
\]
Substituting expression in (\ref{positiveiffgeneral}), we get
\[
(1-p)\alpha{\tau \choose \theta}(\frac{1-\sigma}{2})^\theta(\frac{1+\sigma}{2})^{\tau-\theta}<
p(1-\alpha){\tau \choose \tau-\theta}(\frac{1-\sigma}{2})^{\tau-\theta}(\frac{1+\sigma}{2})^{\theta}.
\]
Rearranging, we get
\[
(\frac{1-\sigma}{1+\sigma})^{2\theta}<(\frac{1-\sigma}{1+\sigma})^{\tau}(\frac{p}{1-p})(\frac{1-\alpha}{\alpha}).
\]
Applying $\log$ on both sides gets us
\[
2\theta\log(\frac{1-\sigma}{1+\sigma})<\tau\log(\frac{1-\sigma}{1+\sigma})+\log(\frac{p}{1-p})+\log(\frac{1-\alpha}{\alpha}).
\]
Solving for $\theta$, we find that the inequality holds if
\[
\theta>\frac{\tau\log(\frac{1-\sigma}{1+\sigma})+\log(\frac{p}{1-p})+\log(\frac{1-\alpha}{\alpha})}{2\log(\frac{1-\sigma}{1+\sigma})}=\left\lceil \frac{1}{2}(\tau-\frac{\log(\frac{1}{p}-1)+\log(\frac{1}{\alpha}-1)}{\log(1+\frac{2\sigma}{1-\sigma})})\right \rceil
\]

For $\theta\geq \left \lceil \frac{1}{2}(\tau-\frac{\log(\frac{1}{p}-1)+log(\frac{1}{\alpha}-1)}{\log(1+\frac{2\sigma}{1-\sigma})})\right\rceil$,
we have $$(1-p)\alpha\Pr[S_{\tau}=\theta|y=0]<
p(1-\alpha)\Pr[S_{\tau}=\theta|y=+1],$$ 
and for $\theta\leq \left \lceil \frac{1}{2}(\tau-\frac{\log(\frac{1}{p}-1)+\log(\frac{1}{\alpha}-1)}{\log(1+\frac{2\sigma}{1-\sigma})})\right \rceil$, we have $$\alpha(1-p)\Pr[S_{\tau}=\theta|y=0]>
(1-\alpha)p\Pr[S_{\tau}=\theta|y=+1].$$ This implies that the maximum is
$\theta^*_{p,\alpha}=\left \lceil \frac{1}{2}(\tau-\frac{\log(\frac{1}{p}-1)+\log(\frac{1}{\alpha}-1)}{\log(1+\frac{2\sigma}{1-\sigma})})\right \rceil$.

\end{proof}

\begin{proof}[Proof of Theorem \ref{thm:majorityLowerBound}]
We start with a skilled candidate. 
The expected number of tests 
that a skilled candidate passes is $\E[S_{\tau}|y=+1]=\tau(\frac{1+\sigma}{2})>\frac{\tau}{2}$.

By using Hoeffding's inequality for Bernoulli distributions, for every $\epsilon>0$,
\[
\Pr[\E[S_{\tau}]-S_{\tau}\geq \epsilon|y=+1]=\Pr[\tau(\frac{1+\sigma}{2})-S_{\tau}\geq \epsilon|y=+1]\leq e^{-2\epsilon^2 \tau}<\delta.
\]
Choosing $\epsilon=\frac{\sigma}{2}$ yields $S_\tau\leq \frac{\tau}{2}<\ceil{\frac{\tau}{2}}$ (as $\tau$ is odd), which holds iff a majority threshold policy would predict that this is an unskilled candidate (false negative). Solving for $\tau$, we get $\tau>\frac{1}{\sigma^2}\ln(\frac{1}{\delta})$.

We now repeat the process for an unskilled candidate. The expected number of tests that an unskilled candidate passes is $\E[S_{\tau}|y=0]=\tau(\frac{1-\sigma}{2})<\frac{\tau}{2}$.

By using Hoeffding's inequality again, we have
\[
\Pr[S_{\tau}-\E[S_{\tau}]\geq \epsilon|y=0]=\Pr[S_{\tau}-\tau(\frac{1-\sigma}{2})\geq \epsilon|y=0]\leq e^{-2\epsilon^2 \tau} <\delta
\]
Choosing $\epsilon=\frac{\sigma}{2}$ yields $S_\tau>\frac{\tau}{2}$, 
which holds iff a majority threshold falsely predicts 
that this is a skilled candidate (false positive). 
Solving for $\tau$ again, we get
$\tau>\frac{1}{\sigma^2}\ln(\frac{1}{\delta})$.\\
Overall, $\tau>\frac{\alpha(1-p)}{\sigma^2}\ln(\frac{1}{\delta})+\frac{p(1-\alpha)}{\sigma^2}\ln(\frac{1}{\delta})=\Omega(\frac{\alpha+p-2p\alpha}{\sigma^2}\ln(\frac{1}{\delta}))$
\end{proof} 
\begin{proof}[Proof of Theorem \ref{thm:minFPwithBudget}]
Let $\pi'$ be any optimal policy for (\ref{optimization1}) (not necessarily threshold) with a fixed number of tests, $\tau$. 
We will show, in two steps, how to transform it 
into an optimal randomized threshold policy. 
The first step is to symmetrize $\pi'$. 
Let $r_k=\Pr[\pi(\hat{y})=1|S_\tau=k]$. 
Define a policy $\pi''$, which performs $\tau$ tests, 
and accepts with probability $r_k$ where $k=S_\tau$. 
Clearly, 
both $\pi'$ and $\pi''$ have the same accept probability. 
In addition, since condition on $S_\tau=k$, 
any sequence of outcomes is equally likely. 
Furthermore, and the probability that $y=1$ 
given any sequence of outcomes with $S_\tau=k$, is identical. (Technically, $S_\tau$ is a sufficient statistics.) 
This implies that the false discovery rate is also unchanged.

This yields that $\pi$ with the randomization vector $r$ is also optimal.

The second step is to suppose---for sake of contradiction---that $\pi''$ 
is not a randomized threshold policy. 
We will show that we can improve the FDR of $\pi''$ 
while keeping the probability of acceptance unchanged. 
This will contradict the hypothesis that $\pi'$ is optimal. 

If $\pi''$ is not a randomized threshold policy, 
then there is no $\theta$ and $k$, 
such that 
\[
r_k=\Pr[\pi(\hat{y}_{i,1},\dots,\hat{y}_{i,\tau}) =1|S_\tau=k\ne \theta]=
 \begin{cases}
       0, &\quad\text{if } k<\theta \\
       1 &\quad\text{if } k>\theta\\
     \end{cases}.
\] 
Now, let $k$ be the minimal value such that $r_k>0$ 
and let $0<i<\tau-k$ be the minimal value 
for which  $0<r_{k+i}<1$.
Clearly, the FDR is lower at $S_\tau=k+i$ than at $S_\tau=k$.
Intuitively, we can shift some probability mass, $\epsilon_k>0$ from $r_k$ to $r_{k+i}$ 
in a way that maintains the acceptance probability of $\pi$ and decreases the false positive rates.

Let $\epsilon_{k+i}>0$ be such that 
$\epsilon_k\cdot r_k=\epsilon_{k+i}\cdot r_{k+i}$.
Let $r'$ be a modified randomization vector for $\pi$ 
such that $r_k'=r_k(1-\epsilon_k), r_{k+i}'= r_{k+i}(1+\epsilon_{k+i})$ 
and for every $l\notin \{k,k+i\}$ $r_l'=r_l$. 
Since $\Pr[\pi(\hat{y}_{i,1},\dots,\hat{y}_{i,\tau}) =1]=\sum_{l=1}^\tau r_l=\sum_{l\notin \{k,k+i\}} r_l+r_k'+r_{k+i}'$, 
the acceptance probability remains the same. 
As for the false discovery rate, since
$\Pr[y_i=0|S_\tau=k+i]<\Pr[y_i=0|S_\tau=k]$, 
$\Pr[S_\tau=k+i]$ is higher with $r'$ than with $r$, $\Pr[S_\tau=k]$ is lower with $r'$  than with $r$ 
and for any $l\notin \{k,k+i\}$, 
$\Pr[S_\tau=l]$ with $r'$ is the same as with $r$, 
the false discovery rate with $r'$ is lower, 
which contradicts the optimality of $\pi$ 
with $r$ as the randomization vector.
\end{proof}

\begin{proof}[Proof of Theorem \ref{thm:posteriorBetter}]

Using Bayes' theorem, the conditional probability can be decomposed as
\[
\Pr[y_i=+1|S_{\tau}=\theta]=\frac{\Pr[y_i=+1]\Pr[S_{\tau}=\theta|y_i=+1]}{\Pr[S_{\tau}=\theta]}=\]
\[
\frac{  p{\tau \choose \theta}(\frac{1-\sigma}{2})^{\tau-\theta}(\frac{1+\sigma}{2})^\theta}
{ p{\tau \choose \theta}(\frac{1-\sigma}{2})^{\tau-\theta}(\frac{1+\sigma}{2})^\theta+
 (1-p){\tau \choose \tau-\theta}(\frac{1+\sigma}{2})^{\tau-\theta}(\frac{1-\sigma}{2})^\theta}.
 \]
 Since $\tau-\theta<\theta$ and ${\tau \choose \theta}={\tau \choose \tau-\theta}$, we get
 \[
\frac{p(1+\sigma)^{2\theta-\tau}}
{ p(1+\sigma)^{2\theta-\tau}+
(1-p)(1-\sigma)^{2\theta-\tau}}=
\frac{p(\frac{1+\sigma}{1-\sigma})^{2\theta-\tau}}
{ p(\frac{1+\sigma}{1-\sigma})^{2\theta-\tau}+
1-p}.
\]
Since $(\frac{1+\sigma}{1-\sigma})>1$ it holds that $(\frac{1+\sigma}{1-\sigma})^{2\theta-\tau}>1$, 
\[
(\frac{1+\sigma}{1-\sigma})^{2\theta-\tau}(1-p)>
1-p.
\]
So,
\[
(\frac{1+\sigma}{1-\sigma})^{2\theta-\tau}>p(\frac{1+\sigma}{1-\sigma})^{2\theta-\tau}+
1-p,
\]
And finally,
\[
\Pr[y_{i'}=+1]=p<\frac{p(\frac{1+\sigma}{1-\sigma})^{2\theta-\tau}}
{ p(\frac{1+\sigma}{1-\sigma})^{2\theta-\tau}+
1-p}=\Pr[y_i=+1|S_{\tau}=\theta].
\]
\end{proof}

\begin{proof}[Proof of Lemma \ref{lemma:beta}]
Let $S'_{\tau}=\sum_{j=1}^\tau (2\hat{y}_{i,j}-1)$, and let $s_{\tau}\in\{-\tau,\dots,\tau\}$ be any of the possible values of $S'_\tau$. Note that
\[
\frac{\Pr[\hat{y}_{i,j}=1|y_i=1]}{\Pr[\hat{y}_{i,j}=1|y_i=0]}=\frac{1+\sigma}{1-\sigma}.
\]
Since the $\hat{y}_{i,j}$  are i.i.d., we have
\begin{align*}
X_{\tau}=&X_0+\sum_{j=1}^\tau(2\hat{y}_{i,j}-1)\cdot
    \log(\frac{\Pr[\hat{y}_{i,j}=+1|y_i=+1]}{\Pr[\hat{y}_{i,j}=+1|y_i=0]})\\
= &\log(\frac{p}{1-p})+S_{\tau}\log(\frac{1+\sigma}{1-\sigma})\\
%
=& \log((\frac{p}{1-p}) (\frac{1+\sigma}{1-\sigma})^{S_{\tau}}).
\end{align*}
Since
\[
\frac{\Pr[S_{\tau}=s_{\tau}|y_i=1]}{\Pr[S_{\tau}=s_{\tau}|y_i=0]}=(\frac{1+\sigma}{1-\sigma})^{s_\tau},
\]
we have
\begin{equation}\label{eq:Xtau}
X_\tau= \log((\frac{p}{1-p}) (\frac{\Pr[S_{\tau}=s_{\tau}|y_i=1]}{\Pr[S_{\tau}=s_{\tau}|y_i=0]})).
\end{equation}
Since 
\[
\Pr[S_{\tau}=s_{\tau}|y_i=1]=\frac{\Pr[S_{\tau}=s_{\tau}]\cdot \Pr[y_i=1|S_{\tau}=s_{\tau}]}{\Pr[y_i=1]}
\]
and 
\[
\Pr[S_{\tau}=s_{\tau}|y_i=0]=\frac{\Pr[S_{\tau}=s_{\tau}]\cdot \Pr[y_i=0|S_{\tau}=s_{\tau}]}{\Pr[y_i=0]},
\]
assigning $\Pr[y_i=0]=1-p$ and $\Pr[y_i=1]=p$, 
we get
\begin{equation}\label{eq:Odds}
\frac{\Pr[S_{\tau}=s_{\tau}|y_i=1]}{\Pr[S_{\tau}=s_{\tau}|y_i=0]}=
\frac{(1-p)\cdot\Pr[y_i=1|S_{\tau}=s_{\tau}]}{p\cdot\Pr[y_i=0|S_{\tau}=s_{\tau}]}.
\end{equation}
Applying (\ref{eq:Odds}) in (\ref{eq:Xtau}) and adding $X_\tau\geq\log \beta$ gives us
\[
X_\tau= \log \left(\frac{\Pr[y_i=1|S_{\tau}=s_{\tau}]}{\Pr[y_i=0|S_{\tau}=s_{\tau}]}\right)=\log\left(\frac{\Pr[y_i=1|S_{\tau}=s_{\tau}]}{1-\Pr[y_i=1|S_{\tau}=s_{\tau}]}\right)\geq \log \beta
\]
\[
\frac{\Pr[y_i=1|S_{\tau}=s_{\tau}]}{1-\Pr[y_i=1|S_{\tau}=s_{\tau}]}\geq  \beta
\]
\[
\Pr[y_i=1|S_{\tau}=s_{\tau}]\geq  \beta(1-\Pr[y_i=1|S_{\tau}=s_{\tau}])
\]
\[
\Pr[y_i=1|S_{\tau}=s_{\tau}]\geq  \frac{\beta}{1+\beta}
\]

Applying (\ref{eq:Odds}) in (\ref{eq:Xtau}) and adding $X_\tau<\log \beta'$ gives us
\[
\frac{\Pr[y_i=1|S_{\tau}=s_{\tau}]}{1-\Pr[y_i=1|S_{\tau}=s_{\tau}]}<  \beta'
\]
Hence
\[
\Pr[y_i=1|S_{\tau}=s_{\tau}]< \frac{\beta'}{1+\beta'}
\]
\end{proof}

\begin{proof}[Proof of Theorem \ref{thm:expectedNumOfTests}]
First recall that given a skilled candidate, for every test $j$,
\[
\Pr[\hat{y}_{i,j}=+1|y_i=+1]=\frac{1+\sigma}{2}
\]
\[
\Pr[\hat{y}_{i,j}=0|y_i=+1]=\frac{1-\sigma}{2}
\]
Hence
\[
\Pr[\hat{y}_{i,j}=0|y_i=1]-\Pr[\hat{y}_{i,j}=+1|y_i=1]=-\sigma.
\]
The lower absorbing barrier is reached 
when a candidate's posterior skill level is lower than the prior of the skill level, i.e.,
\[
\log \frac{\epsilon'}{1-\epsilon'}-\log \left(\frac{1+\sigma}{1-\sigma}\right)
\]
and the starting point is just one step away from the lower absorbing barrier: 
\[
X_0=\log \frac{p}{1-p}.
\]
According to Corollary \ref{cor:piGreedy}, the upper absorbing barrier is in
\[
\log(\frac{1-\epsilon}{\epsilon}).
\]
To derive the results for the expected duration of the random walk for skilled and unskilled candidates, we shift the locations of the absorbing points so that the lower barrier would be in 0 and also divide them by a step size (so now we have that every step is of size $1$). The new upper absorbing barrier is at 
\[
a=\left\lceil\frac{\log(\frac{1-\epsilon}{\epsilon})-
(\log \frac{\epsilon'}{1-\epsilon'}-\log(\frac{1+\sigma}{1-\sigma}))}{\log(\frac{1+\sigma}{1-\sigma})}\right\rceil=
\left\lceil\frac{\log(\frac{(1-\epsilon)(1-\epsilon')(1+\sigma)}{\epsilon \epsilon'(1-\sigma)})}{\log(\frac{1+\sigma}{1-\sigma})}\right\rceil.
\]
And we also shift the starting point:
\[
z=\left\lceil\frac{\log \frac{p}{1-p}-
(\log \frac{ \epsilon'}{1- \epsilon'}-\log(\frac{1+\sigma}{1-\sigma}))}{\log(\frac{1+\sigma}{1-\sigma})}\right\rceil=
\left\lceil\frac{\log (\frac{p(1-\epsilon')(1+\sigma)}{\epsilon'(1-p)(1-\sigma)})}{\log(\frac{1+\sigma}{1-\sigma})}\right\rceil
\]

As stated in \cite{feller1}, the expected duration of a random walk with absorbing barriers of $0$ and $a$ from $z=1$ is (equation 3.4, chapter XIV [page 348]):
\[
\E[\tau_{s}]=\E[D_{z=1}]=\frac{1}{q-p}\left(z-a\cdot\frac{1-(\frac{q}{p})^{z}}{1-(\frac{q}{p})^{a}}\right)=
\frac{1}{-\sigma}\left(z-a\cdot\frac{1-(\frac{1-\sigma}{1+\sigma})^z}{1-(\frac{1-\sigma}{1+\sigma})^{a}}\right).\]
Hence,
\[
\E[\tau_{s}]=\frac{1}{\sigma}\left(a\cdot\frac{1-(\frac{1-\sigma}{1+\sigma})^{z}}{1-(\frac{1-\sigma}{1+\sigma})^{a}}-z\right).
\]
As for unskilled candidates, the absorbing points and the starting point are the same, the only difference is that \[
\Pr[\hat{y}_{i,j}=+1|y_i]=\frac{1-\sigma}{2}
\]
and
\[
\Pr[\hat{y}_{i,j}=0|y_i=+1]=\frac{1+\sigma}{2}.
\]
Therefore,
\[
\Pr[\hat{y}_{i,j}=0|y_i=0]-\Pr[\hat{y}_{i,j}=+1|y_i=0]=\sigma 
\]
and we deduce
\[
\E[\tau_{u}]=\frac{1}{\sigma}\left(z-a\cdot\frac{1-(\frac{1+\sigma}{1-\sigma})^{z}}{1-(\frac{1+\sigma}{1-\sigma})^{a}}\right).
\]
\end{proof}

\begin{proof}[Deviations for the confusion matrix (Table \ref{tab:confMat})]
We split the claim in the confusion matrix (Table \ref{tab:confMat}) into two parts.
First, using equation (2.4) from chapter XIV [page 345] in \cite{feller1}, we get
\[
\text{FNR}=\Pr[\pi_{\text{Greedy}}(\hat{y}_{i,1},\dots,\hat{y}_{i,\tau})=0|y_i=+1]=\frac{(\frac{1-\sigma}{1+\sigma})^a-(\frac{1-\sigma}{1+\sigma})^z}{(\frac{1-\sigma}{1+\sigma})^a-1}
\]
and
\[
\text{TNR}=\Pr[\pi_{\text{Greedy}}(\hat{y}_{i,1},\dots,\hat{y}_{i,\tau})=0|y_i=0]=\frac{(\frac{1+\sigma}{1-\sigma})^a-(\frac{1+\sigma}{1-\sigma})^z}{(\frac{1+\sigma}{1-\sigma})^a-1}.
\]

The second part follows from the fact the gambler's ruin must end in case of absorbing barriers. 
\[
\text{TPR}=\Pr[\pi_{\text{Greedy}}(\hat{y}_{i,1},\dots,\hat{y}_{i,\tau})=1|y_i=+1]=1-\frac{(\frac{1-\sigma}{1+\sigma})^a-(\frac{1-\sigma}{1+\sigma})^z}{(\frac{1-\sigma}{1+\sigma})^a-1}=\]
\[
\frac{(\frac{1-\sigma}{1+\sigma})^z-1}{(\frac{1-\sigma}{1+\sigma})^a-1}=
\frac{\frac{\epsilon'(1-p)(1-\sigma)}
{p(1-\epsilon')(1+\sigma)}-1}
{\frac{\epsilon'\epsilon(1-\sigma)}{(1-\epsilon')(1-\epsilon)(1+\sigma)}-1}=
\frac{\frac{\mu(1-p)}{p}-1}
{\frac{\epsilon\mu}{(1-\epsilon)}-1}=
\frac{(1-\epsilon)(\mu(1-p)-p)}
{p(\epsilon\mu-(1-\epsilon))},
\]
Where $\mu:=\frac{\epsilon'(1-\sigma)}{(1-\epsilon')(1+\sigma)}$. For $\epsilon\leq 1/4$ and $p< 1/2$ we get $0\leq\mu \leq 1/3$ and $\mu=\Theta(\epsilon'(1-\sigma))$, therefore
\[
\text{TPR}=\Theta \left(\frac{p-\mu}{p}\right)=
\Theta\left(1-\frac{\epsilon'}{p}(1-\sigma)\right).
\]

Hence $\text{FNR}=\Theta(\frac{\epsilon'}{p}(1-\sigma))$.

\[
\text{FPR}=\Pr[\pi_{\text{Greedy}}(\hat{y}_{i,1},\dots,\hat{y}_{i,\tau})=1|y_i=0]=\frac{(\frac{1+\sigma}{1-\sigma})^{z}-1}{(\frac{1+\sigma}{1-\sigma})^{a}-1}=
\frac{\frac{p(1-\epsilon')(1+\sigma)}{(1-p)\epsilon'(1-\sigma)}-1}
{\frac{(1-\epsilon')(1-\epsilon)(1+\sigma)}{\epsilon'\epsilon(1-\sigma)}-1}=\]
\[
=
\frac{\frac{p}{(1-p)\mu}-1}
{\frac{(1-\epsilon)}{\epsilon \mu}-1}\frac{\epsilon(p-(1-p)\mu)}{(1-p)(1-\epsilon-\epsilon\mu)}
=\Theta\left(\epsilon(p-\mu) \right)
=\Theta\left(\epsilon(p-\epsilon'+\epsilon'\sigma) \right)
\]
Hence $\text{TNR}=\Theta\left(1-\epsilon(p-\epsilon'+\epsilon'\sigma) \right)$.
\end{proof}

\begin{proof}[Proof of Theorem \ref{thm:expectedNumOfTestsUntilDecision}]

\[
\E[\tau]=\E[\tau_{s}]p+\E[\tau_{u}](1-p)=
\]

\[
=\frac{1}{\sigma}\left(a\cdot\frac{1-(\frac{1-\sigma}{1+\sigma})^{z}}{1-(\frac{1-\sigma}{1+\sigma})^{a}}-z\right)p+
\frac{1}{\sigma}\left(z-a\cdot\frac{1-(\frac{1+\sigma}{1-\sigma})^{z}}{1-(\frac{1+\sigma}{1-\sigma})^{a}}\right)(1-p)=
\]
\[
\approx\frac{1}{\sigma}\left(a\cdot(1-\frac{\epsilon'}{p}(1-\sigma))-z\right)p+
\frac{1}{\sigma}(z-a(\epsilon(p-\epsilon'+\epsilon'\sigma)))(1-p)\approx \frac{ap}{\sigma}
\]
\end{proof}

\subsection{Proofs from Section \ref{sec:twoGroups}}
\label{app:twoGroups}
The next lemma aids in the proof of Theorem (\ref{thm:stochasticDominance}).
\begin{lemma}\label{lemma:binLR}
Let $Z_n^{\eta}$ be a Binomial random variable with parameters $n\in \mathbb{N}$ and $\eta\in(0,1)$. Given a number of successes, $k\in\{0,\dots,n\}$, we know that the probability mass function of $Z_n^{\eta}$ is $f_k(\eta):=\Pr[Z_n^{\eta}=k]={n\choose k}\eta^k(1-\eta)^{n-k}$. Let $\mathcal{L}(\eta|k)$ be the likelihood function of the event $Z_n^{\eta}=k$. Then the maximum likelihood of $f_k(\eta)$ is $\eta=\frac{k}{n}$. I.e., 
\[
\mathcal{L}(\eta|k)=
\text{argmax}_{\eta}f_k(\eta)=
\text{argmax}_{\eta}{n\choose k}\eta^k(1-\eta)^{n-k}=\frac{k}{n}.
\]
\end{lemma}

\begin{proof}[Proof of Lemma \ref{lemma:binLR}]

We notice that ${n\choose k}$ does not depend on $\eta$, thus
\[
\text{argmax}_{\eta}f_k(\eta)=
\text{argmax}_{\eta}{n\choose k}\eta^k(1-\eta)^{n-k}=\text{argmax}_{\eta}\eta^k(1-\eta)^{n-k}\]
The log-likelihood is particularly convenient for maximum likelihood estimation. Logarithms are strictly increasing functions, as a result, maximizing the likelihood is equivalent to maximizing the log-likelihood, i.e.,
\[
\text{argmax}_{\eta}\eta^k(1-\eta)^{n-k}=\text{argmax}_{\eta}\ln(\eta^k(1-\eta)^{n-k})=\text{argmax}_{\eta}k\ln(\eta)+(n-k)\ln(1-\eta)
\]
Differentiating (with respect to $\eta$) and comparing to zero we get
\[
\frac{d\ln(f_k(\eta))}{d\eta}=\frac{k}{\eta}-\frac{n-k}{1-\eta}=0.
\]
And after refactoring,
\[
k(1-\eta)=(n-k)\eta
\]
The function $\ln(f_k(\eta))$ is a strictly concave as its second derivative is negative,
\[
\frac{d^2\ln(f_k(\eta))}{d\eta^2}=-\frac{k}{\eta^2}-\frac{n-k}{(1-\eta)^2}<0,
\]
And since the derivative of a strictly concave function is zero at $\frac{k}{n}$, then $\hat{\eta}=\frac{k}{n}$ is a global maximum.
Therefore, $\hat{\eta}=\frac{k}{n}$ obtains absolute maximum in $f_k(\eta)$.
\end{proof}

\begin{proof}[Proof of Theorem \ref{thm:stochasticDominance}]
Let $Z_\tau^{\eta_i}$ be a random variable 
that represents the number of flips out of a $\tau$-tests sequence 
with a noise level of $\eta_i$,
i.e., $Z_\tau^{\eta_i}$ is the number of times 
when $y_j\ne y$ for $1\leq j \leq \tau$. 
We use $Z_\tau^{\eta_i}$ to express $\Pr[\pi(\hat{y}_{i,1},\dots,\hat{y}_{i,\tau})=1|y_i=0,\eta=\eta_i]$ as the probability that at least $\theta$ flips,
\[
\Pr[\pi(\hat{y}_{i,1},\dots,\hat{y}_{i,\tau})=1|y_i=0,\eta=\eta_i]=
\Pr[Z_\tau^{\eta_i}\geq \theta]
\]
and the probability of $\Pr[\pi(\hat{y}_{i,1},\dots,\hat{y}_{i,\tau})=1|q=+1,\eta=\eta_i]$ as at most $\tau-\theta$ flips, thus 
\[
\Pr[\pi(\hat{y}_{i,1},\dots,\hat{y}_{i,\tau})=1|y_i=+1,\eta=\eta_i]=\Pr[Z_\tau^{\eta_i}\leq \tau- \theta].
\]
From Lemma (\ref{lemma:binLR}) and since  probability density function (pdf) are is monotone increasing, 
we derive that the pdf of $Z_n^{\eta_2}$ 
satisfies \emph{monotone likelihood ratio property} 
over the pdf of $Z_n^{\eta_1}$. 
This implies that the pdf of $Z_n^{\eta_2}$ 
also has \emph{first-order stochastic dominance} over $Z_n^{\eta_1}$ 
by Theorem 1.1 in \cite{Whitt1980Uniform}. 
From \emph{stochastic dominance}, 
we can derive the desired inequalities 
\begin{equation*}
FP_{\theta,\tau}^{\eta_1}=\Pr[\theta\leq Z_n^{\eta_1}]<\Pr[ \theta \leq Z_n^{\eta_2}]
=FP_{\theta,\tau}^{\eta_2}
\end{equation*}
and
\begin{equation*}
FN_{\theta,\tau}^{\eta_1}=\Pr[Z_n^{\eta_1}\leq \tau- \theta]<\Pr[Z_n^{\eta_2}\leq \tau- \theta]=FN_{\theta,\tau}^{\eta_2}.
\end{equation*}
\end{proof}

\section{Gaussian Worker Screening Model Extension}
\label{app:gaussianExtended}
In this extension, we characterize the conditional expectation, $\mathbbm{E}[Q|Y_1,...,Y_n]$ and
 the conditional variance of $Q$ given the tests$Y_i$, i.e., $\text{Var}[Q|Y_1,...,Y_n]$.
 
First, note that because $P(Q)$ is Gaussian, 
and the conditionals $P(Y_i|Q)$ are all Gaussian,
the joint probability $P(Q, Y_1, ..., Y_n)$
is a multivariate Gaussian.
We work out the precise analytic forms
for the mean and variance of the conditional 
$P(Q | Y_1, ..., Y_n)$ in terms of the quality and noise variances 
($\sigma_Q^2$ and $\sigma_\eta^2$) and the number of tests $n$. 

To begin, we note the properties of the joint distribution over $P(Q, Y_1, ..., Y_n)$.
Owing to the generative process for our $Y_i$,
all have mean $\mu_Q$, 
and thus the joint over the means is an $n+1$-dimensional vector
$(\mu_Q, ..., \mu_Q)$.
The full $n+1 \times n+1$ covariance matrix $\Sigma$ has the form
\begin{equation}
\label{eq:covariance}
\Sigma =
\begin{bmatrix}
    \sigma_Q^2 & \sigma_Q^2 &  \dots  & \sigma_Q^2 \\
    \sigma_Q^2 & \sigma_Q^2 + \sigma_\eta^2 & \dots  & \sigma_Q^2 \\
    \vdots & \vdots & \ddots & \vdots \\
    \sigma_Q^2 & \sigma_Q^2 &  \dots  & \sigma_Q^2 + \sigma_\eta^2 
\end{bmatrix}
\end{equation}
where all off-diagonal entries $\Sigma_{ij}$ for $i\neq j$ have value $\sigma_Q^2$ and all diagonal entries $i=j \geq 2$ corresponding to the variance of tests $Y_i$ take value $\sigma_Q^2 + \sigma_\eta^2$. The top-left entry corresponds to the variance of the test Q and thus has variance $\sigma_Q^2$.

We can now derive the equations for the conditional mean and conditional variance of $Q|Y_1,...,Y_n$.
To begin, note the following basic facts about deriving conditionals from multivariate Gaussians:
to estimate the conditional of one set of variables, 
given another set $P(\mathbf{x}_1|\mathbf{x}_2)$ 
we can segment our data matrix into those rows corresponding 
to the variables we don't condition upon (here, just $Q$) 
and those we do (here, $Y_1,..,Y_n$), 
expressing our covariance matrix in terms of the following submatrices:
\[
\Sigma =
\begin{bmatrix}
    \Sigma_{11} & \Sigma_{12} \\
    \Sigma_{21} & \Sigma_{22} \\
\end{bmatrix}
\]
Here, $\Sigma_{11}\in \mathbbm{R}^{1\times 1}$,
$\Sigma_{21}\in \mathbbm{R}^{n \times 1}$,
$\Sigma_{21}\in \mathbbm{R}^{1 \times n}$, and
$\Sigma_{22}\in \mathbbm{R}^{n \times n}$.

The conditional expectation $\mathbbm{E}[Q|Y_1,...,Y_n]$
is then expressed as 
\begin{equation} 
\label{eq:mean}
\mathbbm{E}[Q|Y_1,...,Y_n] =
\mu_Q + \Sigma_{12} \Sigma_{22}^{-1} (\mathbf{y}-\mu_y)
\end{equation}
and the conditional variance $Var[Q|Y_1,...,Y_n]$
can be expressed as
\begin{equation}
Var[Q|Y_1,...,Y_n] = \Sigma_{11}
- \Sigma_{12} \Sigma_{22}^{-1} \Sigma_{21}
\end{equation}
which should be familiar as the Schur complement
of the $n \times n$ matrix $\Sigma_{22}$.
Intuitively, this corresponds to inverting the full matrix $\Sigma$, deleting those rows and columns corresponding to observed variables, and then inverting the resulting ($1 \times 1$) matrix back.

What remains is to show that for the particular 
covariance matrix that interests us (Equation \ref{eq:covariance}), these expressions have simple analytically computable forms. 
Specifically we state the following simple theorems.

\begin{theorem}
\label{thm:expectation}
For jointly Gaussian variables $Q, Y_1, ..., Y_n$
characterized by the covariance matrix given in (Equation \ref{eq:covariance}),
the conditional expectation takes form \begin{equation}
\mathbbm{E}[Q|Y_1,...,Y_n]
= \mu_Q + \left[\frac{1}{\frac{\sigma_\eta^2}{\sigma_Q^2} + n}, \dots \right] \cdot (\mathbf{y} - \mathbf{\mu}_y)
\end{equation}
\end{theorem}

\begin{theorem}\label{thm:variance}
For jointly Gaussian variables $Q, Y_1, ..., Y_n$
characterized by the covariance matrix given in (Equation \ref{eq:covariance}),
the conditional variance of $Q$ given the tests $Y_i$ takes form \begin{equation}
\text{Var}[Q|Y_1,...,Y_n] =
\frac{1}{
\frac{1}{\sigma_Q^2} + \frac{n}{\sigma_\eta^2}}
\end{equation}
\end{theorem}

\subsection{Proofs from Appendix \ref{app:gaussianExtended}}
\label{app:gaussian}
\begin{proof}[Proof of Theorem \ref{thm:expectation}]

The crucial step to apply Equation \ref{eq:mean}
to this data is to work out a simple expression 
for the inverse of the $n\times n$ submatrix $\Sigma_{22}$.
We recall that this matrix is symmetrical with all diagonal entries equal to $\sigma_Q^2 + \sigma_\eta^2$ and all off diagonals equal to $\sigma_Q^2$.

We call upon a lemma due to \cite{miller1981inverse}.
Which states that when $A$ is invertible and $B$ is a rank-1 matrix, the inverse of their sum takes the following form:

\begin{equation}
(A+B)^{-1} = A^{-1} - \frac{1}{1 + \text{Trace}(B A^{-1})} A^{-1} B A^{-1} 
\end{equation}

We can decompose $\Sigma_{22}$
into such an $A$ and $B$
by defining $B$ to be the matrix
that takes value $\sigma_Q^2$ everywhere
and $A$ to be a diagonal matrix that takes values
$\sigma_\eta^2$ (along the main diagonal). 
Thus $\Sigma_{22}^{-1}= (A+B)^{-1}$ and we can proceed by applying the lemma.

First we note that $A^{-1}$ is a diagonal matrix 
with all entries on the main diagonal set to 
$\frac{1}{\sigma_\eta^2}$.
Then we note that $B A^{-1}$ 
is an $n \times n$ matrix with all entries 
set to $\frac{\sigma_Q^2}{\sigma_\eta^2}$.
Thus, $\text{Trace}(B A^{-1}) = \frac{n \sigma_Q^2}{\sigma_\eta^2}$.

The matrix $A^{-1} B A^{-1}$ has all entries equal to $\frac{\sigma_Q^2}{\sigma_\eta^4}$,
and thus our desired inverse $\Sigma_{22}^{-1}$ 
can be expressed as follows:
$$\Sigma_{22}^{-1} = (A+B)^{-1} 
=
\begin{bmatrix}
    \frac{1}{\sigma_\eta^2}  &    &0 \\
     & \ddots &   \\
    0  & & \frac{1}{\sigma_\eta^2}\\
\end{bmatrix}
- 
\frac{1}{
1 + \frac{n \sigma_Q^2}{ \sigma_\eta^2}}
\begin{bmatrix}
    \frac{\sigma_Q^2}{\sigma_\eta^4}  &  \dots & \dots \\
    \vdots & \ddots & \vdots  \\
    \vdots  & \dots & \ddots \\
\end{bmatrix}
$$
Thus each off-diagonal entry takes values
\begin{align}
-\frac{\frac{\sigma_Q^2}{\sigma_\eta^4}}{
1 + \frac{n \sigma_Q^2}{ \sigma_\eta^2}} &=
%
%
-\frac{\frac{\sigma_Q^2}{\sigma_\eta^4}}{
1 + \frac{n \sigma_Q^2}{ \sigma_\eta^2}} 
\cdot \frac{\frac{\sigma_\eta^2}{\sigma_Q^2}}
{\frac{\sigma_\eta^2}{\sigma_Q^2}}\\
& = -\frac{\frac{1}{\sigma_\eta^2}}{
\frac{\sigma_\eta^2}{\sigma_Q^2}+ n}\\
& = -\frac{1}{
\frac{\sigma_\eta^4}{\sigma_Q^2}+ n \sigma_\eta^2}
\end{align}
and each on-diagonal entry has an additional $\frac{1}{\sigma_\eta^2}$ term that comes from $A^{-1}$.
\begin{equation}
\frac{1}{\sigma_\eta^2}
- \frac{1}{
\frac{\sigma_\eta^4}{\sigma_Q^2}+ n \sigma_\eta^2}
\end{equation}

Now that we know the precise expression for all entries of $\Sigma_{22}^{-1}$,
we can calculate the vector-matrix product 
$\Sigma_{12} \Sigma_{22}^{-1}$.
Because every entry of $\Sigma_{12}$
takes value $\sigma_Q^2$, and because 
every column of $\Sigma_{22}^{-1}$ has the same $n$ values (just in different order),
the product $\Sigma_{12} \Sigma_{22}^{-1}$
is an $n$ dimensional vector, where
all $n$ values are equal:
\begin{align}
\Sigma_{12} \Sigma_{22}^{-1} & = 
\left[
\frac{\sigma_Q^2}{\sigma_\eta^2}
- \frac{n \sigma_Q^2}{
\frac{\sigma_\eta^4}{\sigma^2} + n \sigma_\eta^2
},
\dots
\right] \nonumber \\
 & = \left[
    \frac{\frac{\sigma_Q^2}{\sigma_\eta^2}
    \left(
    \frac{\sigma_\eta^4}{\sigma_Q^2} + n \sigma_\eta^2 \right) n\sigma_Q^2
    }
    {\frac{\sigma_\eta^4}{\sigma_Q^2} + n \sigma_\eta^2},
\dots
\right]\nonumber \\
 & = \left[
    \frac{\sigma_\eta^2 }
    {\frac{\sigma_\eta^4}{\sigma_Q^2} + n \sigma_\eta^2},
\dots
\right] \nonumber \\
 & = \left[
    \frac{1}
    {\frac{\sigma_\eta^2}{\sigma_Q^2} + n },
\dots
\right]  \label{eq:quality-mean}
\end{align}
This expression for $\Sigma_{12}\Sigma_{22}^{-1}$,
together with the definition of the conditional expectation 
of a multivariate Gaussian (Equation \ref{eq:mean}) concludes the proof.

\end{proof}

\begin{proof}[Proof of Theorem \ref{thm:variance}]

We can now produce the expression for 
$\Sigma_{12} \Sigma_{22}^{-1} \Sigma_{21}$.
Because every entry of the $1 \times n$ matrix $\Sigma_{12} \Sigma_{22}^{-1}$
takes value
$$\frac{1}{\frac{\sigma_\eta^2}{\sigma_Q^2} + n }$$
and because every entry in the $n \times 1$ matrix $\Sigma_{21}$
takes value $\sigma_Q^2$, 
\begin{equation}
\Sigma_{12} \Sigma_{22}^{-1} \Sigma_{21} =
\frac{n \sigma_Q^2}{\frac{\sigma_\eta^2}{\sigma_Q^2} + n }
\end{equation}
The expression for the conditional variance
$\text{Var}[Q|Y_1,...,Y_n]$ follows:

\begin{align}
\text{Var}[Q|Y_1,...,Y_n] & = \Sigma_{11} - \Sigma_{12} \Sigma_{22}^{-1} \Sigma_{21} \nonumber \\ 
&=
\sigma_Q^2 - \frac{n \sigma_Q^2}{\frac{\sigma_\eta^2}{\sigma_Q^2} + n }
\nonumber \\
& = 
\frac{\sigma_Q^2 \left( \frac{\sigma_\eta^2}{\sigma_Q^2} + n \right) 
- n \sigma_Q^2}{
\frac{\sigma_\eta^2}{\sigma_Q^2} + n
} \nonumber \\
& = 
\frac{\sigma_\eta^2}{
\frac{\sigma_\eta^2}{\sigma_Q^2} + n
} \nonumber \\
& = 
\frac{1}{
\frac{1}{\sigma_Q^2} + \frac{n}{\sigma_\eta^2}
} \label{eq:quality-variance}
\end{align}
\end{proof}

\subsection{Proofs}
\label{app:gaussianTwoGroups}
\begin{proof}[Proof of Theorem \ref{thm:EqVariances}]
First, recall that
\[
\text{Var}[Q|Y_1,...,Y_n] =
\frac{1}{\frac{1}{\sigma_Q^2} + \frac{n}{\sigma_\eta^2}}=
\frac{\sigma_Q^2\sigma_\eta^2}{\sigma_\eta^2+n\sigma_Q^2}.
\]
Solving for $n_2$ in the equation $\text{Var}_1[Q|Y_1,...,Y_{n_1}] =\text{Var}_2[Q|Y_1,...,Y_{n_2}]$,

\[
\frac{\sigma_Q^2\sigma_{\eta_1}^2}{\sigma_{\eta_1}^2+n_1\sigma_Q^2}=
\frac{\sigma_Q^2\sigma_{\eta_2}^2}{\sigma_{\eta_2}^2+n_2\sigma_Q^2}
\]
we get
\[
\sigma_{\eta_1}^2(\sigma_{\eta_2}^2+n_2\sigma_Q^2)=
\sigma_{\eta_2}^2(\sigma_{\eta_1}^2+n_1\sigma_Q^2)
\]
and hence
\[
\sigma_{\eta_1}^2n_2=
\sigma_{\eta_2}^2n_1.
\]
Extracting $n_2$, we find that $n_2=\frac{\sigma_{\eta_2}^2}{\sigma_{\eta_1}^2}n_1$.
\end{proof}

\begin{proof}[Proof of Theorem \ref{thm:EqExpectations}]
First, recall that
\[
\mathbbm{E}[Q|Y_1,...,Y_n]
= \mu_Q + \left[\frac{1}{\frac{\sigma_\eta^2}{\sigma_Q^2} + n}, \dots \right] \cdot (\mathbf{y} - \mathbf{\mu}_y)=\mu_Q + \left[\frac{\sigma_Q^2}{\sigma_\eta^2 + n\sigma_Q^2}, \dots \right] \cdot (\mathbf{y} - \mathbf{\mu}_y)
\]
Now,
\[
\mathbbm{E}_1[Q|Y_1,...,Y_{n_1}]-\mathbbm{E}_2[Q|Y_1,...,Y_{n_2}]=
\]
\[
 \left[\frac{\sigma_Q^2}{\sigma_{\eta_1}^2 + n_1\sigma_Q^2}, \dots \right] \cdot (\mathbf{y_1} - \mathbf{\mu}_y)-
 \left[\frac{\sigma_Q^2}{\sigma_{\eta_2}^2 + n_2\sigma_Q^2}, \dots \right] \cdot (\mathbf{y_2} - \mathbf{\mu}_y)
\]
\[
=\frac{\sigma_Q^2}{\sigma_{\eta_1}^2 + n_1\sigma_Q^2}n_1(\mathbf{\bar{y_1}})-
\frac{\sigma_Q^2}{\sigma_{\eta_2}^2 +n_2\sigma_Q^2}n_2(\mathbf{\bar{y_2}})
\]
\[
=\frac{\sigma_Q^2n_1}{\sigma_{\eta_1}^2 + n_1\sigma_Q^2}(\mathbf{\bar{y_1}})-
\frac{\sigma_Q^2n_2}{\sigma_{\eta_2}^2 +n_2\sigma_Q^2}(\mathbf{\bar{y_2}})
\]

\[
=\frac{\sigma_Q^2n_1}{\sigma_{\eta_1}^2 + n_1\sigma_Q^2}(\mathbf{\bar{y_1}})-
\frac{\sigma_Q^2\frac{\sigma_{\eta_2}^2}{\sigma_{\eta_1}^2}n_1}{\sigma_{\eta_2}^2 +\frac{\sigma_{\eta_2}^2}{\sigma_{\eta_1}^2}n_1\sigma_Q^2}(\mathbf{\bar{y_2}})
\]
\[
=\frac{\sigma_Q^2n_1}{\sigma_{\eta_1}^2 + n_1\sigma_Q^2}(\mathbf{\bar{y_1}})-
\frac{\sigma_Q^2n_1}{\sigma_{\eta_1}^2 +n_1\sigma_Q^2}(\mathbf{\bar{y_2}})
\]
\end{proof}

\end{document}